# Impact of Robot Facial-Audio Expressions on Human Robot Trust Dynamics and Trust Repair

Hossein Naderi[1], Alireza Shojaei[2], Philip Agee[3], Kereshmeh Afsari[4], Abiola Akanmu[5]

[1]PhD Candidate, Dept. of Building Construction, Myers-Lawson School of Construction, Virginia Tech, Blacksburg, VA. ORCID: https://orcid.org/0000-0002-6625-1326. Email: hnaderi@vt.edu

[2]Assistant Professor, Dept. of Building Construction, Myers-Lawson School of Construction, Virginia Tech, Blacksburg, VA (corresponding author). ORCID: https://orcid.org/0000-0003-3970-0541. Email: shojaei@vt.edu

[3]Assistant Professor, Dept. of Building Construction, Myers-Lawson School of Construction, Virginia Tech, Blacksburg, VA, email: pragee@vt.edu

[4]Assistant Professor, Dept. of Construction Engineering and Management, Myers-Lawson School of Construction, Virginia Tech, Blacksburg, VA, email: keresh@vt.edu

[5]Associate Professor, Dept. of Construction Engineering and Management, Myers-Lawson School of Construction, Virginia Tech, Blacksburg, VA, email: abiola@vt.edu

**Abstract**

Despite recent advances in robotics and human-robot collaboration in the AEC industry, trust has mostly been treated as a static factor, with little guidance on how it changes across events during collaboration. This paper investigates how a robot's task performance and its expressive responses after outcomes shape the dynamics of human trust over time. To this end, we designed a controlled within-subjects study with two construction-inspired tasks, Material Delivery (physical assistance) and Information Gathering (perceptual assistance), and measured trust repeatedly (four times per task) using the 14-item Trust Perception Scale for HRI plus a redelegation choice. The robot produced two multimodal expressions, a "glad" display with a brief confirmation after success, and a "sad" display with an apology and a request for a second chance after failure. The study was conducted in a lab environment with 30 participants and a quadruped platform, and we evaluated trust dynamics and repair across both tasks. Results show that robot success reliably increases trust, failure causes sharp drops, and apology-based expressions partially restores trust (44% recovery in Material Delivery; 38% in Information Gathering). Item-level analysis indicates that recovered trust was driven mostly by interaction and communication factors, with competence recovering partially and autonomy aspects changing least. Additionally, age group and prior attitudes moderated trust dynamics with younger participants showed larger but shorter-lived changes, mid-20s participants exhibited the most durable repair, and older participants showed most conservative dynamics. This work provides a foundation for future efforts that adapt repair strategies to task demands and user profiles to support safe, productive adoption of robots on construction sites.

1. Introduction

Robotics has been emerging rapidly in the Architecture, Engineering, and Construction (AEC) industry, with recent reviews documenting accelerating progress in autonomous mobile robots, manipulation, and on-site collaboration as companies try to enhance productivity, address labor shortages, and improve safety outcomes [1]. These trends suggest that future construction workflows will increasingly integrate robots alongside human workers, rather than replace them [2]. At the same time, construction sites are open, dynamic, and safety-critical, uncertainty and task variability make human-in-the-loop decision making essential for managing the tasks and workflows. Consequently, human-robot interaction (HRI) becomes central for two reasons: first, workers can calibrate when and how to rely on robotic assistance; second, site efficiency and acceptance depend on smooth, intelligible interaction between workers and robots [3], [4]. Within HRI, trust is an important factor of whether people choose to rely on a robot and how they recover from failures over time. Theoretical work shows that trust shapes reliance and compliance, and that lack of trust can lead to misuse (over-reliance) or disuse (under-reliance) with real safety consequences in high-risk environments like construction settings [5], [6]. Expressive behaviors, such as facial expressions, displays of regret, or apologies, can influence perceived competence, and responsibility by users, and they have been reported to affect trust formation and trust repair after errors [7], [8], [9].

Within construction industry, several studies have investigated the trust-building between workers and robots [10], [11], [12], [13], However, most of them regarded the trust as a static concept, while it is actually dynamic [14]. To manage dynamic nature of trust, it is essential to understand trust calibration, which can be defined as a process where workers continuously adjust their trust in robots based on their observed performance and reliability over time. Trust calibration ensures that workers' trust aligns with the actual capabilities of the robot, preventing overtrust or undertrust, both of which can be detrimental to productivity of construction tasks in future human-robot collaboration [15]. For example, overtrust can lead workers to rely too much on robots, even in situations where the robot may not be fully capable, potentially leading to errors or accidents [16]. On the other hand, undertrust which is a result of trust violation can result in workers not using the robot to its full potential, reducing the efficiency and benefits that robots can offer [17]. Therefore, dynamic calibration of trust through continuous feedback and mechanisms from robots is crucial to fostering an appropriate level of trust in construction environments.

Moreover, trust repair and trust dampening are essential mechanisms in managing the dynamic nature of trust in human-robot interactions. Trust repair strategies, such as robots demonstrating improved performance after a failure or providing clear explanations for past mistakes, can help rebuild trust when it has been violated due to errors [18]. Conversely, trust dampening involves mechanisms like transparency about robot limitations or confidence scores to prevent overtrust, ensuring that workers do not overly rely on the robot beyond its actual capabilities [19], [20]. Both approaches are critical for maintaining calibrated trust, which allows workers to use robots effectively without compromising safety or productivity. Despite the importance of this field in future of automated robot-human collaboration, Chang, and et al. [14] introduced this field as a knowledge gap in construction studies and recommended further research. Therefore, the identified knowledge gaps in this paper are listed as follows: (1) Most studies in the construction context regard human trust as a static concept, overlooking the importance of dynamic trust calibration in human-robot collaboration; (2) The impact of robot expressive behavior on the dynamics of trust and trust repair remains an unexplored area.

To address these research gaps, this paper defines three main questions: (1) how do expressive robot behaviors (a "glad" success display and a "sad + apology + request-for-a-second-chance" repair display) shape the dynamics of human-robot trust over time during construction-relevant work? (2) to what extent can those expressions facilitate trust repair following a task failure? (3) which task factors, and participant factors, such as baseline attitudes toward robots, prior experience, and demographics, most strongly moderate trust dynamics and repair? To this end, we conducted a controlled within-subjects experiment with 30 participants performing two construction-inspired tasks, Material Delivery and Information Gathering, while interacting with an expressive mobile robot.

## 2. Background
### 2.1. Trust Definition in HRI

In the context of human-robot interaction, trust can be defined as "the attitude that an agent will help achieve an individual's goals in a situation characterized by uncertainty and vulnerability [3]". There are different factors that affect trust in human robot teaming. These factors can be categorized into three main groups: (1) human related factors; (2) robot-related factors, and (3) environmental factors [21]. First category includes the impact of factors such as gender, nationality, and culture on trust. For example, some studies [22], [23] have found that male participants have higher tendency to trust robots compared to female participants. Other factors relevant to characteristics of individuals such as emotions, and self-confidence also play a significant role in shaping trust. Positive emotions can increase trust, while low self-confidence, especially in unfamiliar contexts, can lead to hesitation in trusting robots [22], [24]. Second category include factors relevant to robot appearance, personality, and communication style. For example, robots with more human-like features can foster higher trust levels [25], although some studies suggest that too much human-likeness can sometimes cause discomfort or distraction [26]. Within this category, performance-based factors, such as transparency, reliability, and competence, directly impact how much users trust robots. For instance, robots that consistently perform tasks accurately and transparently, with clear explanations of their actions, are generally perceived as more trustworthy [27]. The third category involves environmental factors, which include mental workload, task nature, and time pressure. Environments that are complex and demand high mental capacity may reduce trust in robots as users may struggle to monitor or understand robot actions under stress [28]. Similarly, time pressure can affect trust; in high-pressure situations, users might over-rely on robots without fully assessing their capabilities [29].

## 2.2. Theory: Trust in Automation

Moreover, effective human-robot collaboration is shaped by several key factors, including human factors (such as cognitive workload and prior experience), robot-specific factors (like safety and predictability), and interaction dynamics (such as a sense of control and task allocation) [30]. Among these, trust is a critical human-related factor that significantly affects the acceptance and effectiveness of robotic systems. [31]. Therefore, without establishing this fundamental trust in human-robot teams, achieving higher level of automation through deliberative acting theory becomes unattainable. The Trust in Automation (TIA) [3] is a theoretical framework for studying how trust can be established and calibrated between humans and autonomous agents, including robots. In construction, where we hypothesized to achieve higher level of automation via foundation models and deliberative acting theory, trust becomes a critical factor in determining whether humans will rely on robots for essential tasks and assign responsibility with confidence [14], [32]. When trust is well-calibrated, it leads to smoother collaboration, improved task efficiency, and better overall outcomes [33]. However, if trust is lacking or poorly calibrated, it can disrupt collaboration [34]. Humans might hesitate to use robots, doubting their abilities and leading to inefficiencies in overall performance of construction tasks. On the other hand, humans in collaboration might overtrust automated robots, leading to potential failures when robots reach their limitations. As a result, the absence of trust can result in inefficiencies, miscommunication, or even safety hazards, undermining the benefits of automated robots in construction environment.

## 2.3. Relevant Work

Prior studies in construction HRI have examined worker trust, but the recent studies showed most capture trust as a static outcome rather than a trajectory that changes across events [14]. For example, Charalambous et al. developed an industrial human robot trust scale and measured trust across different type of robots, however it does not model and capture how trust dynamic evolves during work sequences (i.e., success, failure, and repair) in construction settings [35]. In applied construction contexts, You et al. proposed the Robot Acceptance Safety Model and showed how design can improve perceived safety, but again reported post-trial attitudes rather than time-varying trust within a task [36]. More recent construction experiments simulated collaborative tasks to study who gets blamed when things go wrong, and workers who attributed failures to themselves retained more trust than those who blamed the robot, an insight about attribution, but not about how trust recovers after a failure [37]. Classic HRI findings also generalize to construction-like risks: faulty navigation reduces people's willingness to comply with a robot's guidance (competence violations depress trust), while emergency-evacuation studies show the opposite hazard, such as people sometimes overtrust and follow unreliable robots under pressure [38], [39]. Therefore, the literature gives factors (reliability, competence, environment) and measurement conditions, but leaves open how workers' trust in construction changes over time across success, failure, and attempted repair, and whether concrete expressive behaviors from robots (e.g., apologies) measurably aid trust repair on job-like tasks.

Robot expressions are the observable signals a robot produces to convey internal state or social intent, signals that people routinely use to judge reliability, intent, and trust in HRI. Visual displays (faces, gaze, LEDs, posture), auditory cues (non-verbal sounds, voice quality, and wording), and motion/proxemics (approach distance, speed, trajectory) are the main modalities typically discussed [40]. Prior work shows that facial expressions can shape perceived trustworthiness and social engagement in HRI and are a practical channel for signaling apology [41]. Gaze is also another form of expression that helps coordinate interaction and influence social judgments under different situations [42]. Moreover, Samarakoon et al. [43] reported proxemics/motion timing systematically affect trust and robot acceptance. On the auditory side, voice naturalness and apologies can raise perceived trust and compliance [44], [45]. Beyond single channels, trust-repair studies emphasize that using different strategies such as apology, denial, promise, and explanation are often effective for violation repairs [46]. Importantly for our design choice, recent studies showed that combining face and voice tends to produce stronger, more reliable trust effects than either alone, because the channels provide redundancy and cross-validation for sincerity [47], [48], [49]. Accordingly, we paired facial expressions with brief spoken voices (see detailed in Figure 3) to signal apology or repair in construction-style tasks. To keep the repair channel practical for construction tasks, we use a brief combination of visual-audio signals rather than lengthy speech, consistent with guidance that this combination can help calibrate trust when the task the situation is unreliable such as construction settings [50]. Therefore, our study is aimed to close human robot trust gaps by (1) measuring dynamic trust in a within-subject sequence of events and (2) testing an expressive repair strategy using a multi modal approach that applies robot face expressions in addition to apology voices. We also compare two construction-inspired task families, **material delivery** (physical assistance) and **information gathering** (perceptual assistance), because task nature are known to shape trust and overtrust, yet are rarely investigated within current studies [14].

3. **Methodology and Experiment Design**

This section describes what methods are used to measure trust to meet the study's objectives. In HRI, trust is typically assessed with three families of methods: (1) self-report methods, which can be defined as standardized questionnaires that ask participants to report their perceived trust (e.g., Jian et al.'s [51] Checklist for Trust in Automation; or Muir & Moray's [52] items; and the Trust Perception Scale for HRI [53]); (2) behavioral indicators are one of the other methods. This method usually includes following observable choices such as delegating a task to the robot, response time, and changes in behavior, which can indicate varying levels of trust based on how users interact with robots and decide during tasks [54]; (3) psychophysiological measures, signals like heart rate, EDA, eye movements, or EEG that can be measured to find the trust level [55]. While promising, these measures can be difficult to interpret due to the complex relationship between physiological states and trust [56]. Self-reports remain the most common and best-documented option in HRI because they are economical, easy to administer repeatedly, and supported by extensive validation and review work [57]. Accordingly, this paper adopted a self-report-centered design in iterative format (not static and one time) and used behavioral indicators as secondary checks. This method is grounded in the measurement literature (narrative and systematic reviews) and in widely used trust scales for automation and HRI [58].

This paper used a two-tier self-report mechanism for measuring trust during the experiments. Figure 1 illustrates three components in two groups of before experiment and after experiment. Once users accepted the participation invitation, they were asked with four questions regarding demographic information, including age, gender, education level, and prior experience with robots. Next, General Attitude towards Robots Scale (GAToRS) were utilized as a validated tool to measure how participants think about the robot before their first experience and impression with autonomous robot [59]. This measurement was deployed with two reasons: (1) this baseline helps interpret between-participant differences during the study and allow to identify the relevant social factors that may impact trust dynamics and trust repair; (2) to capture social attitudes toward robots before working with them to prevent any potential biases. Both of these questionnaires were asked only one time per users. Full details of these questionnaires can be found in Appendix A. In layer B (see Figure 1), we utilized the 14-item version of Trust Perception Scale for HRI [53], which is a validated sub-scale of Schaefer's Trust Perception Scale designed specifically for HRI and intended for repeated measurement over time [60]. Prior reports document this 14-item subscale as suitable for repeated measurement in interactive studies [61]. This scale measured trust at five time points (T0-T4) with the 14-item Trust Perception Scale-HRI (TPS-HRI), developed for HRI and designed to track trust changes.

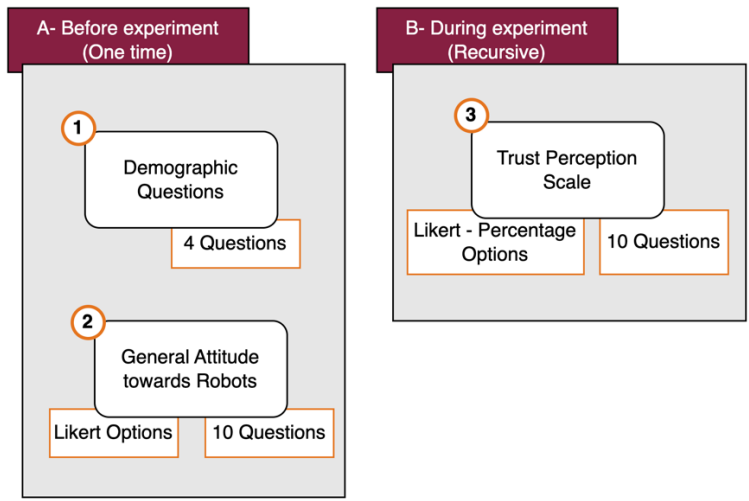

*Figure 1* Overall mechanism of trust measurement

### 3.1. Experiment Tasks and Scenarios

This section describes the scenarios used to measure trust. We selected two tasks to identify differences between tasks and avoid tying results to a single activity, since task characteristics are known to affect trust in automation and HRI [62]. Evaluations based on one task can over- or under-estimate trust depending on the task's demands; testing two tasks with different natures provides a stronger basis for generalization [62]. To represent common job-site interactions, we modeled two following tasks (1) a material delivery task, asking robot for physical assistance where a robot carries objects for a worker (e.g., delivering fasteners, tools, or bricks; staging materials at points of use); and (2) an information gathering task, asking robot for sensing/inspection duties where a robot collects information from the environment (e.g., counting bricks or pallets, progress capture, hazard/condition checks). Both categories are prevalent in construction robotics deployments [63].

In the material delivery task, participants were asked to build a three columns nine rows wall with foam bricks and blocks following a simple specification sheet (Appendix C). Before interacting with the robot, participants inspected the brick and block stock at their workstation and calculated any shortfall relative to the target layout. They then collaborated with the robot via a web application interface (see Figure 2 from 1 to 4A) to request additional units: pressing the "request" button once. The robot's role was physical assistance: navigate to the supply area, pick the requested item, and deliver it to the participant's bench. This sequence created clear opportunities to observe trust-related choices under routine site-like conditions. In the information gathering task, participants directed the robot to obtain and report a site-relevant fact rather than deliver an object. Using the same web application interface (see Figure 2), they commanded the robot to travel to the storage area and determine how many bricks remained after prior work. The robot then navigated to the designated location, performed the query (e.g., visual inspection), and returned the result to the participant through the interface and on-robot display. Participants used the reported number to complete a short digital task, creating a practical dependency on the robot's information quality. This scenario stands in for inspection and situational-awareness duties that are common on job sites, inventory checks, progress verification, and simple condition assessments, allowing us to examine how expressive behavior influences trust when the robot's output is informational rather than physical.

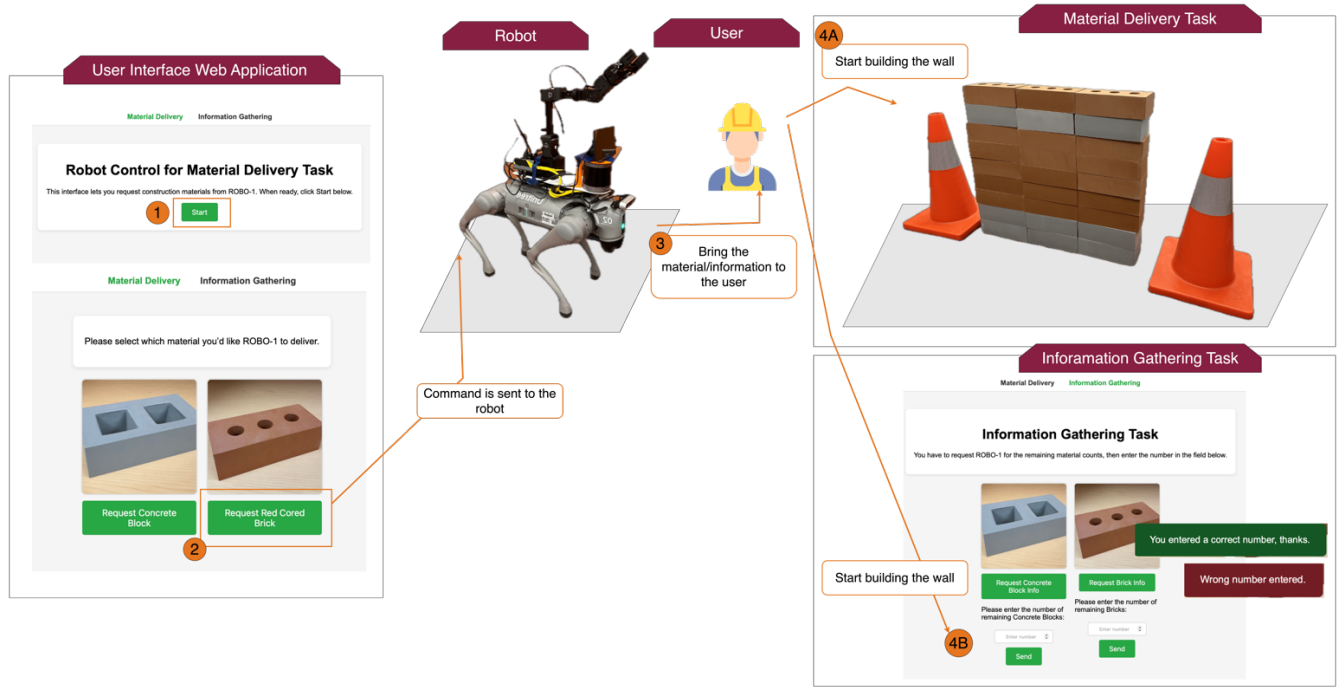

*Figure 2* Experiment workflow and components

Figure 3 illustrates the process of conducting experiment (module B in Figure 3). Participants began with a brief document that described the robot's capabilities and limits (see Appendix C for more details). This document is comparable to a first impression a worker might have before operating a newly purchased robot. This pre-brief established the baseline trust measure prior to any hands-on interaction. During the session, trust was measured at key moments after specific events (see T1-1 to T2-3 in Figure 3) using the 14-item Trust Perception Scale. In addition to these items, users were asked to give a estimate of their overall trust to the whole system after each event, following the order shown in Figure 3. The robot first succeeds on a task and displays a "glad" expression; participants report trust; the robot then fails on a task; participants report trust; the robot attempts repair with a sad face in addition to a scripted apology sentence for a second chance; participants report trust; finally, participants provide an overall trust judgment and indicate whether they would delegate a similar task again. If they agree to continue, the same sequence is repeated on the other task, enabling a within-subject comparison across task types. To control order effects, we randomized task order across participants and counterbalanced where the scripted failure occurred, so that each participant experienced one success episode and one failure across tasks. This design lets us observe trust trajectories within individuals while comparing physical-assistance versus information-gathering contexts, consistent with evidence that task type influences trust formation and recovery.

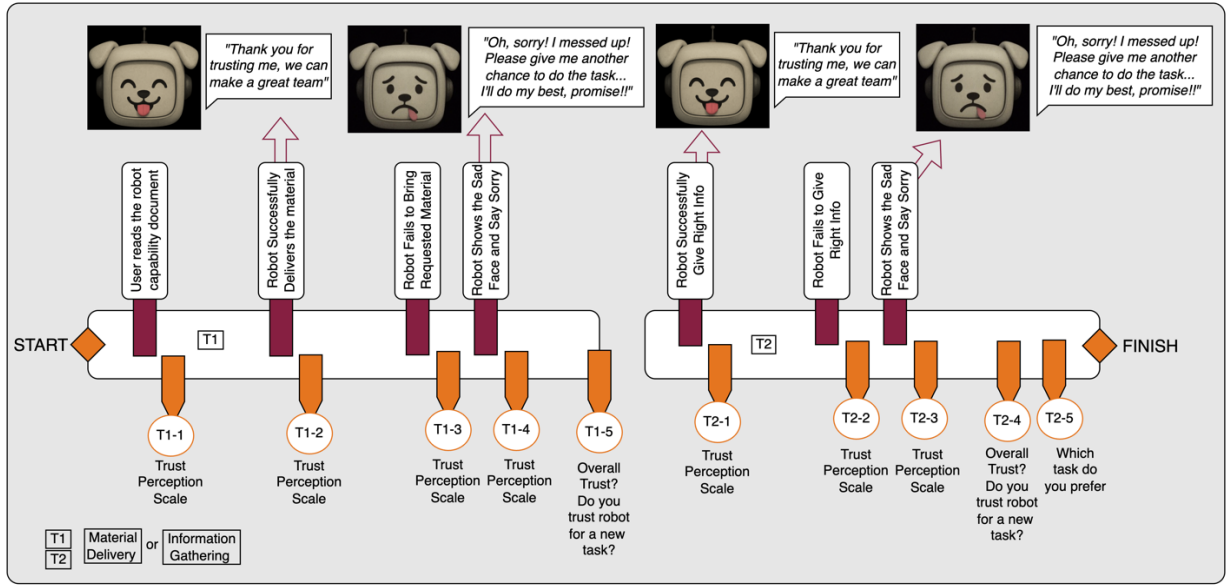

*Figure 3* The process of conducting experiment across two tasks with trust measurement points

### 3.2. Sample size

Because each participant in our study experiences all time points and expressive-behavior events, the design naturally falls under a within-subjects (repeated-measures) approach: the same people are measured repeatedly as the interaction continues, which reduces between-person noise and increases statistical power for detecting change over time. Standard methodologies of repeated measures emphasize these advantages and the need to check repeated-measures assumptions when analyzing a single factor such as time. We analyzed the main effect of time (four trust measurements) in a repeated-measures framework, noting the sphericity assumption (equal variances of all pairwise difference scores) and the use of Greenhouse-Geisser corrections if it is violated [64]. We set α = 0.05 and power = 0.80, use Cohen's f to index ANOVA effect size with f=0.25 (medium), and assume a moderate within-person correlation ρ=0.50 under compound symmetry, consistent with repeated-measures tutorials [65], [66]. These choices align with best-practice workflows and with implementations in G*Power 3.1[67]. Let k=5 time levels with sphericity ε≈1. So, the degrees of freedom are $df_1=(k-1)\varepsilon=4$ and $df_2=(N-1)(k-1)\varepsilon=4(N-1)$. In repeated measures with compound symmetry, the effective error variance is reduced in proportion to 1−ρ, which inflates the noncentrality parameter λ by 1/(1−ρ). Combining this property with Cohen's f, yields the standard approximation according to the equation 1.

$$\lambda = \frac{f^2}{1-\rho} df_2 \qquad f = 0.25, \rho = 0.50 \tag{1}$$

Substituting $f^2$=0.0625 and ρ=0.50 gives λ=0.125 $df_2$, which equals to = 0.125 × 4(N−1). Solving the noncentral-F power problem for α=0.05, 1−β=0.80, and $df_1$=4 yields a required λ≈12.1. Setting 0.5 (N−1)=12.10 gives number of participants (N≈25.2). To accommodate mild sphericity deviations, uncertainty in ρ, and attrition, we plan N=30, which preserves ≥0.80 power under reasonable departures from the assumptions. This choice is consistent with prior studies within the context of trust experiments that detected effects with samples at or below 30, including 26 in Robinette et al work [68] on over-trust in emergency evacuation and 24 in Desai et al. work on changing reliability and trust [69]. Studies using NARS methodology in live HRI also report 28 participants, supporting feasibility in this range [70].

### 3.3. Experiment Implementation

To implement the study, we used a quadruped platform with integrated sensing and expressive hardware. Figure 4 shows the robot and its attachments: a Unitree Go2 Edu base (for mobility), a K1 robotic arm (for pickup/delivery), a RoboSense Helios-32 LiDAR (for localization and obstacle detection), and a small monitor/speaker used for participant-facing expressions and brief spoken messages during success, failure, and repair episodes. Recruitment notices were distributed to Virginia Tech students across departments. Of the 46 students who expressed interest, 30 completed an individual session. Each session lasted 90-120 minutes and followed the sequence described earlier (Section 3.2).

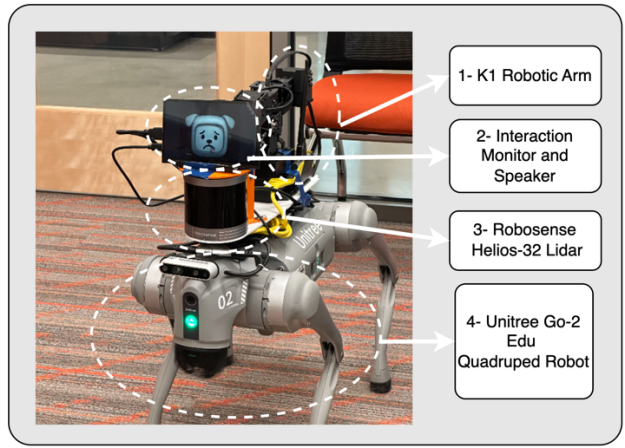

*Figure 4* Robotic hardware and attachments

Figure 5 depicts the laboratory layout used in all sessions. (1) Material storage area included the foam bricks/blocks used in both tasks. (2) Robot path parts A-B indicate the route for deliveries or information reporting to the participant; and parts C-D indicate the return path from the task area to storage. (3) User desk was a rolling workstation with a laptop running the web application for issuing requests to the robot and completing the TPS-HRI questionnaires. (4) Task area was the marked zone where participants assembled the foam-brick wall or waited for receiving the information report. This standardized layout was designed to make sure consistent approach angles and distances for the robot while keeping the participant's interaction flow unchanged across sessions. Moreover, Figure 6 provides images from both tasks. The left panel shows a participant receiving a block from the robot during the Material Delivery task. The right panel shows the same participant during the Information Gathering task as the robot displays an inventory result (e.g., "3" remaining bricks) on its monitor. These images illustrate the typical interaction posture, handoff distance, and the use of the on-robot display/speaker for expressive feedback; participant faces are obscured for privacy.

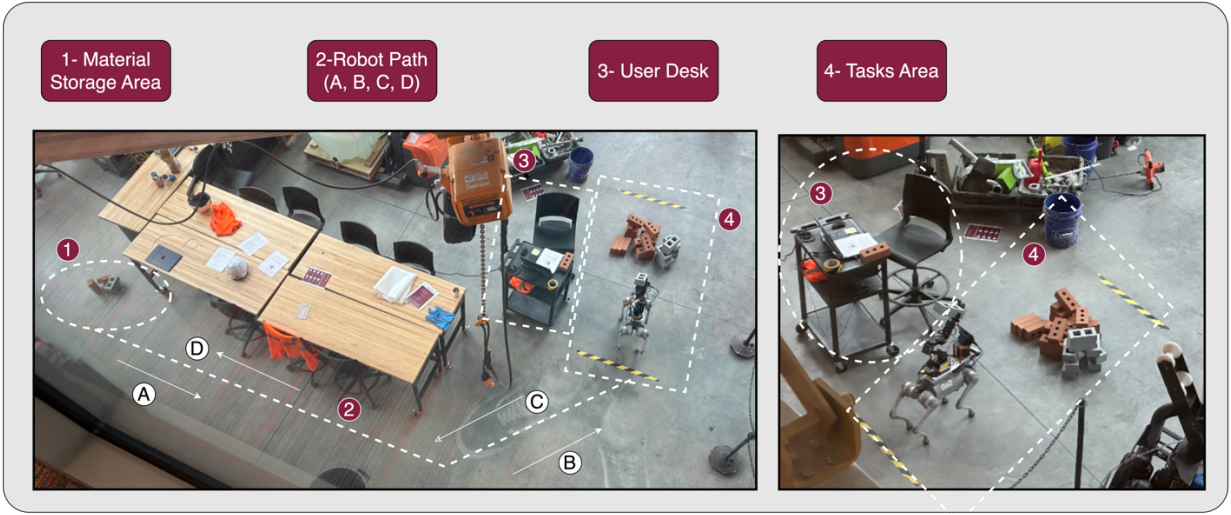

Figure 5 Designed lab layout for the experiment

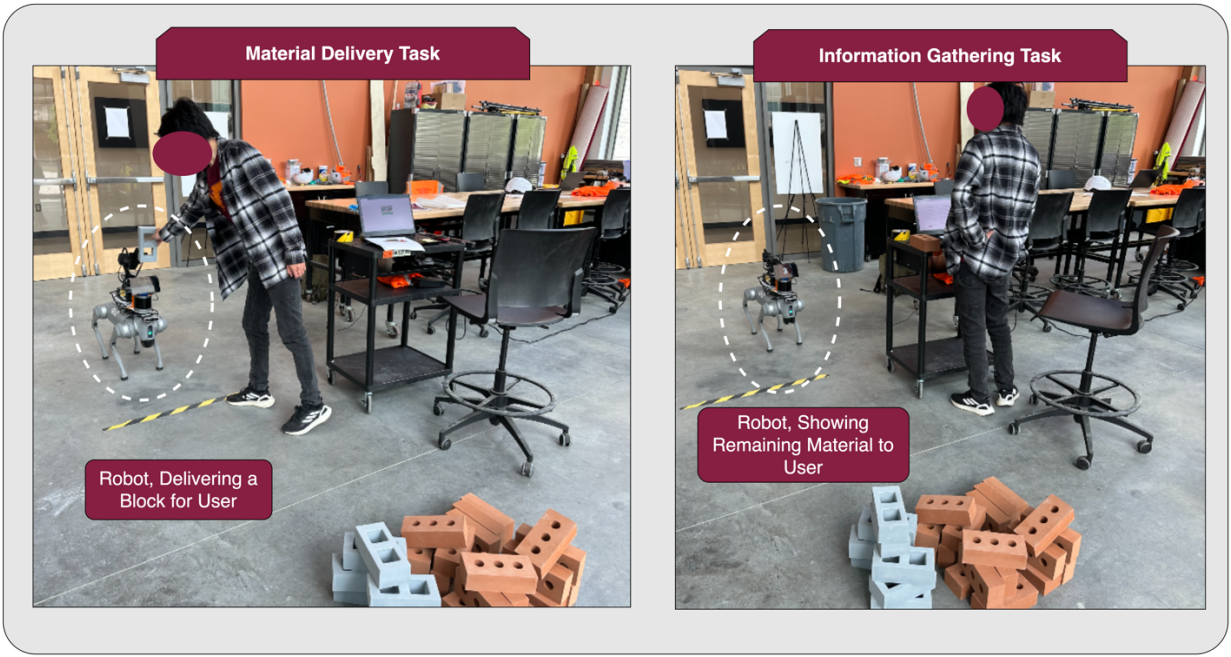

Figure 6 Participant collaboration with the robot over two tasks

## 4. Results
### 4.1. Participant distribution

This subsection explains the distribution of participants in our study to document the sample for reproducibility. Figure 7 illustrates this distribution over three main factors of gender (most left), educational level (middle), and age groups (most right). 23 of participants were men, while the rest 7 were women. For education level, the sample included 16 PhDs, 2 Master, and 12 Bachelor students. The age groups were divided into three categories: 12 participants (40%) in 18-24 year old group, 8 participants (25%) between 25-30 years old, and the remaining 10 participants were categorized on over 30 age group. The average age was 27.1 years (standard deviation = 5.6).

These counts are simple integer conversions of the plotted percentages (minor rounding expected). The mix reflects a research-university setting with a larger share of advanced-degree. These three distributions were reported to describe the pool for the reproductivity and also used in the next sections to analyze different trust patterns.

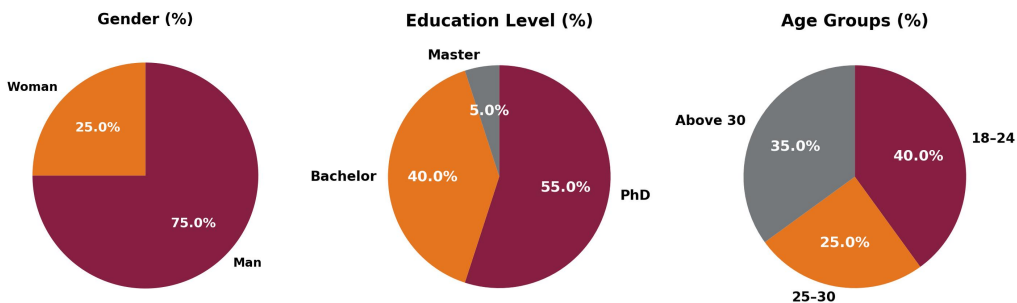

*Figure 7* Participant distribution

### 4.2. Attitude toward robots

Table 1 lists the ten baseline questions we used to measure general attitudes toward robots. We grouped them into two themes: usefulness (Q1-Q3, Q9, Q8) and comfort (Q6, Q7, Q10, Q4 and Q5), which we reverse-coded so that higher values indicate greater comfort or less anxiety. Moreover, Figure 8 shows the item-by-item distributions, sorted roughly from more to less favorable based on the participant's opinion. In usefulness category, the first two items show strong agreement: for Q1 ("robots are necessary for hard/dangerous jobs") about 90% of participants agreed or strongly agreed, and Q2 ("robots can make my life easier") showed a similarly high share. Support is positive but slightly softer for Q3 ("robots are a natural product of our civilization") and Q9 ("professionally supervised robots are safe for construction"), where most participants agreed but more selected neutral compared to Q1-Q2. The most mixed response in this group is Q8 ("robots can be trusted"): roughly a third agreed, a majority chose neutral, and a small minority disagreed. In short, participants broadly endorsed the usefulness of robots and the value of supervision on job sites, while general trust (Q8) was more uncertain factor.

*Table 1* General attitude toward robots' questions and their categories

| ID | Question | Category |
|---|---|---|
| Q1 | Robots are necessary because they can do jobs that are too hard or too dangerous for people. | Usefulness |
| Q2 | Robots can make my life easier | Usefulness |
| Q3 | Robots are a natural product of our civilization. | Usefulness |
| Q4 | Widespread use of robots is going to take away jobs from people | Less anxiety (comfort) |
| Q5 | If I became too dependent on robots, something bad might happen. | Less anxiety (comfort) |
| Q6 | I would feel very nervous just being around a robot. | Less anxiety (comfort) |
| Q7 | Robots scare me. | Less anxiety (comfort) |
| Q8 | Robots can be trusted. | Usefulness |
| Q9 | Professionally supervised robots are safe enough to use in construction tasks. | Usefulness |
| Q10 | I would feel relaxed talking with robots. | Less anxiety (comfort) |

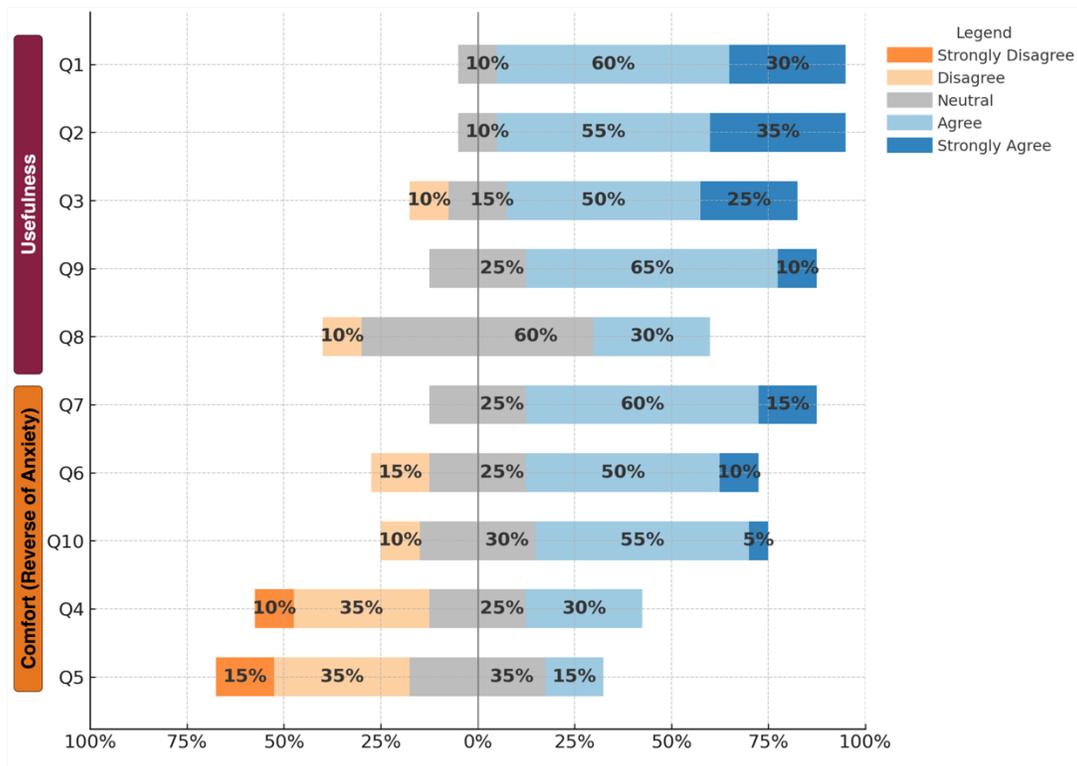

*Figure 8* Ranking and favorability of general attitude over trust factors

After reversing the concern items, Q7 ("robots scare me") indicates high comfort and most participants effectively rejected being afraid. Q10 ("I would feel relaxed talking with robots") is also positive but more moderate, with many choosing neutral. Q6 ("I would feel nervous around a robot") shows the some spread in this group, suggesting that some participants feel nervous even when overall fear is low. The two last questions in the figure highlight the most significant concerns: Q4 (job displacement) is one of the most anxiety areas, with many responses on the negative side of the plot, indicating that job-loss issue is a real concern for them; Q5 (over-dependence risk) shows even a more anxiety, reflecting concerns about relying too much on robots. Therefore, the figure suggests a pattern for our sample: participants see robots as useful and safe when supervised, feel mostly comfortable interacting with them, but remain uncertain about trust in the abstract and concerned about impacts such as job loss and over reliance.

### 4.3. Trust dynamics

Figure 9 aims to demonstrate the dynamic nature of trust in two different construction task by showing how trust changed over time during the study. The x-axis lists nine events, which the first point (T1-1) is the baseline taken before any hands-on interaction, after participants only read a short document about the robot. This is used as the reference level for all later measurements. The next four points (T1-2 to T1-5) cover the Material Delivery task: success with a happy face (T1-2), a scripted task failure (T1-3), the robot's repair attempt with a sad face and apology (T1-4), and a short post-reflection measure (T1-5). The last four points (T2-1 to T2-4) repeat the same sequence for the Information Gathering task. The y-axis reports the mean overall trust score (0-100) across all 30 participants at each event, and the shaded area around the line shows ±1 standard deviation, which indicates how different the individual scores were at each moment. For Material Delivery, trust starts at 65.3 at baseline and rises to 76.5 after the robot completes the task and displays a happy face, which is the highest point in the entire timeline. The subsequent failure drops trust to 57.0, a decrease of 19.5 points from the success level, and even less than the trust baseline. After the repair display (sad face in addition to apology), trust came back to 65.6, showing a 44% trust repair. Following a brief pause for reflection, the final reading for this task is 65.2, slightly lower compared to the immediate post-repair value (down 0.4). Therefore, we can conclude that success boosts trust well above baseline, failure produces a sharp drop, and the apology partially repairs trust to near-baseline levels.

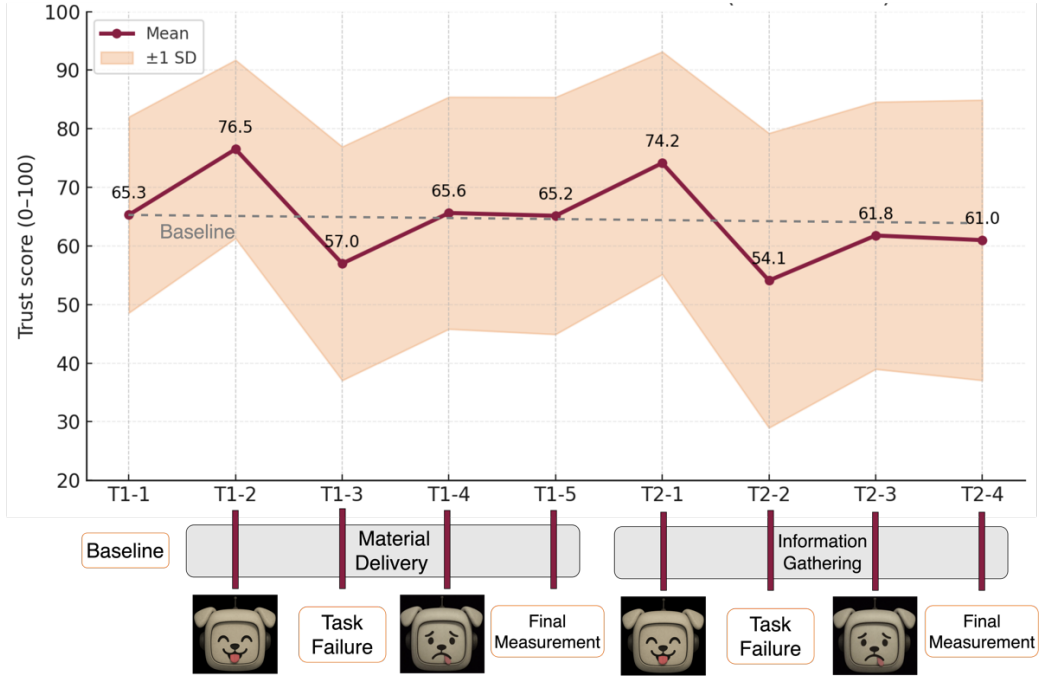

*Figure 9* General trust dynamics over two tasks (mean) plus standard deviation

For Information Gathering, the pattern repeats but with slightly weaker recovery. Trust rises to 74.2 at success, then falls to 54.1 at failure, a 20.1-point drop, which is a larger drop compared to the drop in material delivery task. The repair expressions improve the trust to 61.8, which is a 38% of the trust lost. After some reflections it settles at 61.0. Comparing the two tasks, the gain after success is larger in Material Delivery, the drop after failure is slightly smaller, and the repair is more effective (44% vs. 38%). Looking across the full sequence, trust after two failures ends just below baseline (61.0 vs. 65.3). The wider SD shades near the two failure points indicate greater disagreement among participants at those moments, which we examine further in the discussion.

### 4.4. Trust perception scale factors (item-level dynamics)

To look inside the overall trust scores, we analyzed the 14-item Trust Perception Scale-HRI across key moments in the study. The y-axis in Figure 10 lists a shortened label for each item (e.g., reliable, predictable, communicates effectively). Scores are shown on a 0-100 scale and higher scores are equivalent to more favorable ones. The x-axis contains seven events that mirror the sequence used earlier: baseline and success-failure-repair for Material Delivery (T11-T14), followed by success-failure-repair for Information Gathering (T21-T23). Reading the figure row by row shows how each dimension of trust moved with outcomes and expressions, rather than only the overall composite. There are several notable observations worth mentioning. First, the largest drops after failure occur on "meets task needs", where T12 score of 86 drops to 41 in T13. This is aligned with the same drop in other task which T21's score for 81 drops to 42 in T22. This trend aligns with the nature of experiment that when the robot fails, the belief that it meets the task requirement collapses most.

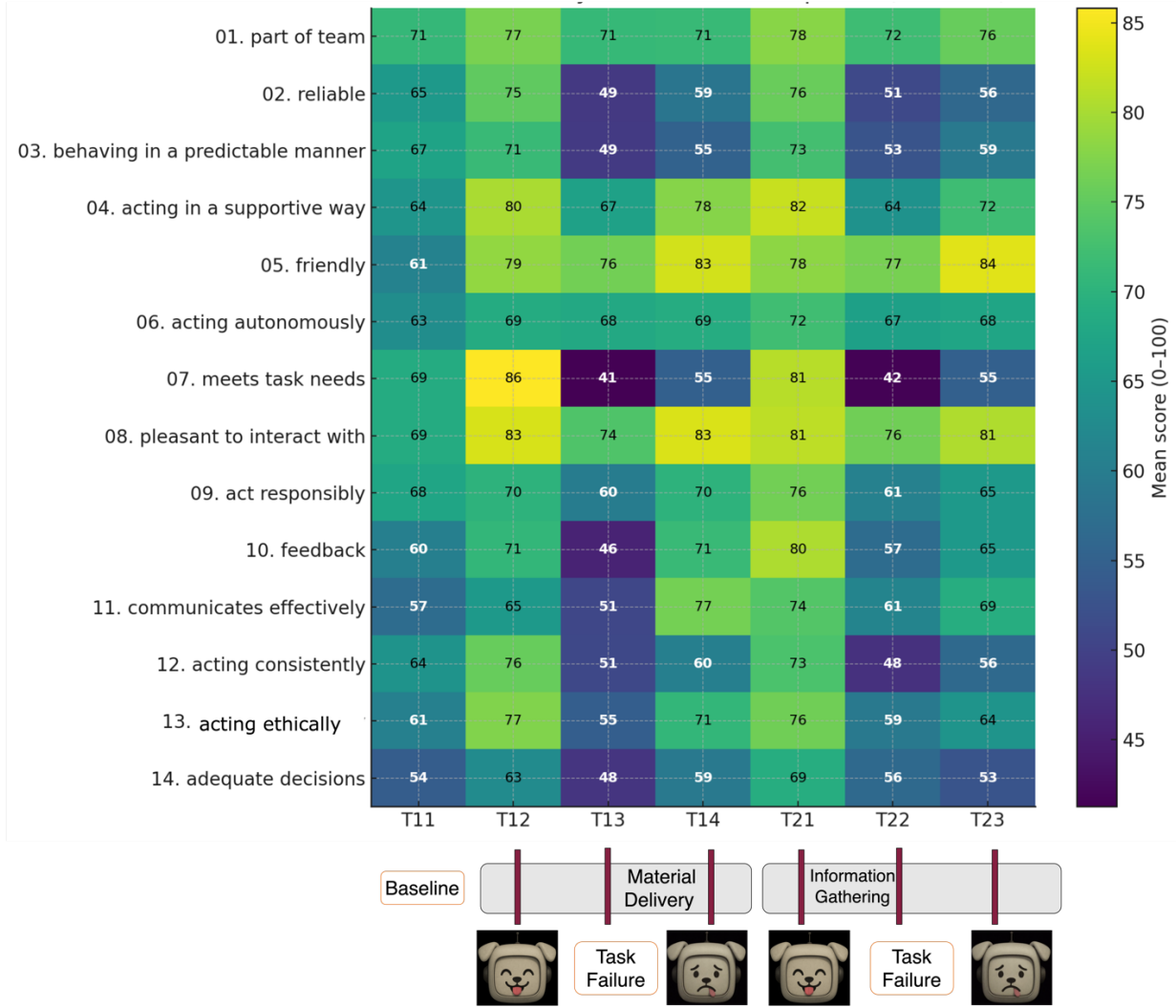

Figure 10 Item level analysis of trust factors and aspects during two task experiments

To make the item-level patterns in the heatmap easier to discuss, we created the composite plot in Figure 11. Each line aggregates conceptually related items on a 0-100 scale: (1) Warmth/Interaction (e.g., friendly, supportive, pleasant to interact with, communicates effectively, feedback); (2) Competence/Consistency (e.g., reliable, predictable, meets task needs, acts consistently); and (3) Autonomy/Decision-making (e.g., acts autonomously, acts responsibly, adequate decisions). The x-axis follows the same sequence of events as before, baseline and the success-failure-repair cycle for Material Delivery, then the same cycle for Information Gathering, so the composite trends can be read alongside the heatmap to check whether category-level movements match item-level changes. First, after the initial success in the material-delivery task, all three dimensions rise, but Warmth/Interaction and Competence/Consistency climb the most, both reaching the 70s, while Autonomy/Decision-making moves up more modestly into the mid-60s. Second, when the scripted failure occurs, the competence lines shows the steepest decline, falling to the low 50s, while warmth reduces to the mid-60s and autonomy remains near the upper-50s, this confirms that performance errors primarily damage perceived capability and reliability rather than interpersonal impressions. Third, the repair expressions (sad face, apology, and request for a second chance) produce strong recovery in Warmth/Interaction, which returns to the high-70s. This shows that the participants agree that the robot has more successful friendly interaction than they expected from baseline point. Regarding the Competence/Consistency, we observed a partial recovery which moves only into the low-60s, and steady improvement in Autonomy/Decision-making, which climbs back to its first position after drop, demonstrating a successful trust repair from autonomy perspective with robot expressions. The same success-failure-repair sequence in the information-gathering task repeats the pattern but with two main differences. First is that the autonomy not only recovered but only shows a slight drop, demonstrating that the nature of information gathering task reflects less autonomy in the perspective of participants when its failure is followed with robot apology expressions. Second, the robot expressions improved the trust repair in both competency and friendliness factors, but sightlier than their Material Delivery counterparts. While we observed a sharp increase in trust repair with robot expressions in Material Delivery (higher than baseline), the friendliness factors were partially repaired. In practical terms, participants continued to see the robot as pleasant and communicative after apologies specially in the Material Delivery task while they remained uncertain about its reliability and autonomy, especially in the Information Gathering task.

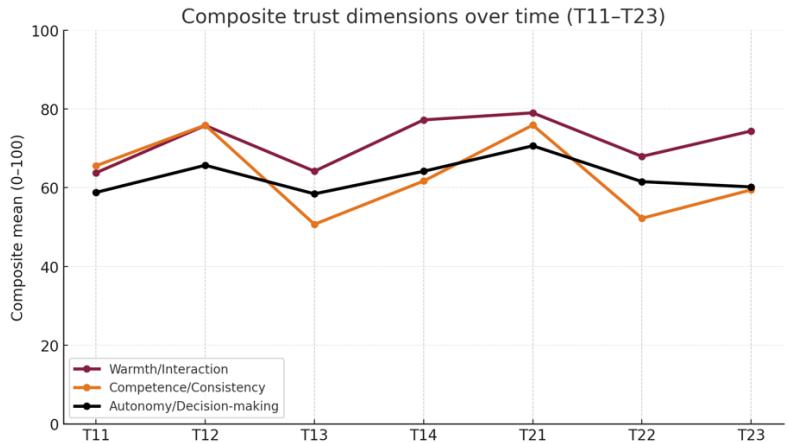

*Figure 11* Composite or categorized trust factor dynamics over two tasks.

## 5. Discussion
### 5.1. Trust Dynamics by Age Group

This section aims to demonstrate the impact of age groups on the trust dynamics in the experiment tasks. Figure 12 shows the mean of trust over time for the three age groups in our sample (see section 4.1) against the overall mean (dashed line) and its ±1 SD band (gray area). The x-axis follows the same nine events as before (baseline; success-failure-repair-reflection for Material Delivery; then the same sequence for Information Gathering). The y-axis shows the 0-100 trust score averaged within each age group at each event. The 18-24 group shows the sharpest changes in both directions. Trust jumps steeply after the happy-face success in each task and then falls again steeply after failures, staying well below the group's own trust baseline. In both tasks the apology increases the trust, but this gain fades quickly (see the post-reflection point in T15 which goes downward again). This trend suggests that expressive behavior for trust repair has a short life for this group. When it comes to comparison of two tasks, their trust is higher for material delivery than for information gathering, suggesting that observable and physical help (bringing a block) is easier to credit than a reported fact (inventory count). Overall, the pattern implies that younger participants are responsive to the robot's expressions, easier to win over with positive expressions, but they are also quick to withdraw trust after errors.

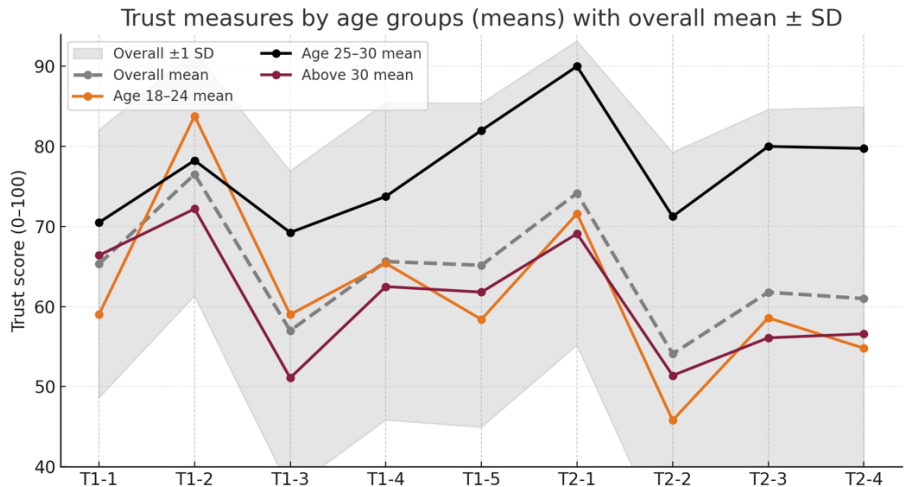

*Figure 12* Trust dynamics (mean) by age groups

Participants above 30 show a more conservative profile. The trust increase after success is smaller than the other groups, as maybe they are holding some reserve about future performance. However, the drops at failure are still strong, showing that evidence of error clearly matters. Repair via the sad face and apology partially restores trust, more noticeably in the Material Delivery task, but the reflection point is almost flat (see T1-4 to T1-5), indicating that the designed pause to reconsider does not move this group much after the apology. In short, older participants appear conservative in both directions: modest increase when things go well, strong penalties when things go wrong. The third group (age 25-30) exhibits the most stable and calibrated pattern. The trust rises after the successful implementation, but when failure occurs their trust falls back toward baseline rather than below it, unlike the younger and older groups. Importantly, the apology and sad face produces a sustained recovery, and trust goes back to or above the baseline level and stays elevated at the reflection points in both tasks. This group also reaches the highest peaks during the study (see point in T2-1). Overall, the age analysis suggests that expressive behavior is most strong but least durable for the youngest group, moderately effective and steady for the oldest group, and effective and durable for participants in the middle group. Practically, this suggests to several design choices: (1) for younger users, more frequent or richer repair expressions may be needed to maintain recovered trust; (2) for older users, clear performance evidence (e.g., explanation, verification) may matter more than additional affect; (3) for the middle group, the present repair strategy appears sufficient if failures are not repeated a lot. These patterns are descriptive rather than causal, but they help explain why the same expressive policy can feel persuasive to some construction workers and not working to others.

### 5.2. Trust Dynamics by General Attitude toward Robots

This subsection examines how prior attitudes toward robots shape trust during the study. To this end, top five optimistic and pessimistic participants based on results in the section 4.2 were put into two main group and plotted their trust trends in Figure 13. The figure shows the same nine-event timeline used earlier on the x-axis. The y-axis reports the 0 to 100 trust score averaged within each group, and the dashed gray line provides the overall mean for reference. For the optimistic group, trust begins high and reacts strongly to successful implementation of material delivery. After the first task's success with the happy-face display, trust climbs steeply and reaches one of the highest values observed in the study. When a failure occurs, the decrease is modest, and the mean remains above the group's own baseline. The repair expression (sad face and apology) and the brief reflection did not shift opinions further, indicating that once these participants have strong prior beliefs that is not impacted by robot's expressions. A similar pattern appears in the information-gathering task, trust rises again at success, drops slightly at failure, and then went downward after the apology rather than up, suggesting that for optimistic participants the verbal apology is neither necessary nor especially persuasive in the absence of new performance evidence, and their prior positive belief carry most of the weight.

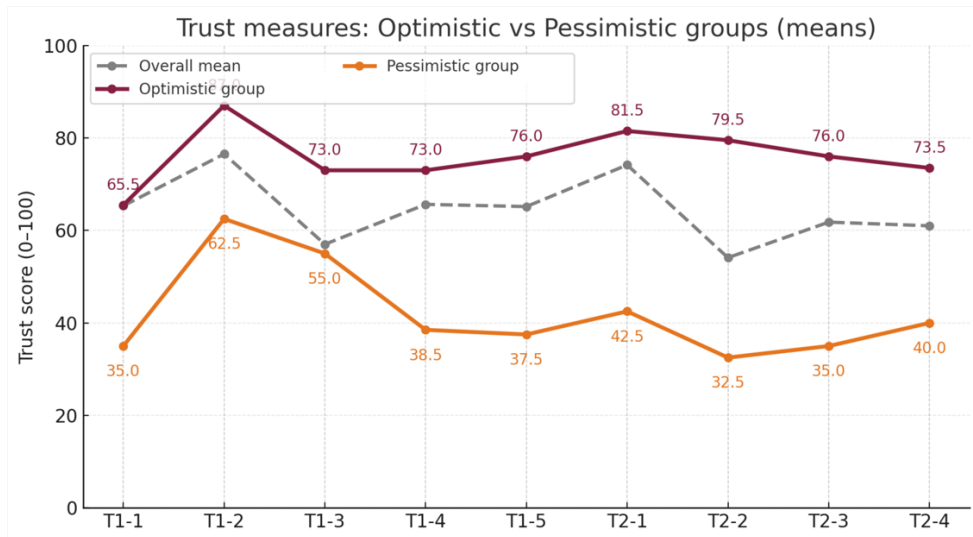

*Figure 13* Trust dynamics by optimistic and pessimistic groups

The pessimistic group shows a different picture, starting from a much lower trust baseline and giving larger penalties to errors while also reacting reversly to the repair. After the first success, trust increases sharply but still below the overall mean (other participants), and after the failure the value went back and the apology even reduced trust further, placing the group close to its starting level by the end of the first task. In the information gathering task, the success increase is smaller, the failure brings trust below baseline, and the apology only recovers to approximately baseline with a small improvement after reflection. This profile suggests that participants with pessimistic ideas put more weight on observable performance rather than on expressive cues, and they tend to treat apologies as time-costly signals rather than corrective action. During think aloud periods in the study, participants gave reasons that align with these curves. Some of the optimistic participants remarked that a dog-like face paired with human-like speech felt inconsistent, and they would have preferred expressions that matched the embodiment (e.g., a non-verbal signal or a short tonal sound) instead of a human-style apology. Several pessimistic participants emphasized that, in construction settings with noise and time pressure, an apology is less valuable than immediate action and they preferred transparent explanations over scripted regret. Taken together, these perspectives point to three design implications: (1) tailor expressive strategies to users' prior beliefs, optimistic users may need more confirmation of continued competence, while pessimistic users benefit from corrective actions instead of expressions; (2) align the modality and style of expression with the robot's morphology (3) pair any apology with concrete corrective action or verification so that the expression serves as a gateway to restored performance rather than a stand-alone social gesture.

### 5.3. Material Delivery vs Information Gathering

This subsection discusses how participants trust about continuing to work with the robot after failures and which task family they would prefer to use the robot for going forward. After each task, participants were asked a simple question: "Would you give the robot another chance to complete a similar task?" The left panel of Figure 14 shows the share who answered Yes. Willingness to allocate another task remained very high after a single failure (90% agreed to give a second chance), but it fell to 63% after two failures, a 27-point drop. Participants who withdrew trust after the second failure commonly noted that "two errors in two simple tasks" suggested the system was not yet reliable enough for the pace and consequences of construction work, however those who still agreed to continue said that an apology and clear acknowledgment of the mistake justified one more try. We also asked participants to choose which task type they would prefer for future deployment, assuming the same level of autonomy and the same expressive repair policy. The right panel of Figure 14 right shows a clear preference for Material Delivery (77%) over Information Gathering (23%).

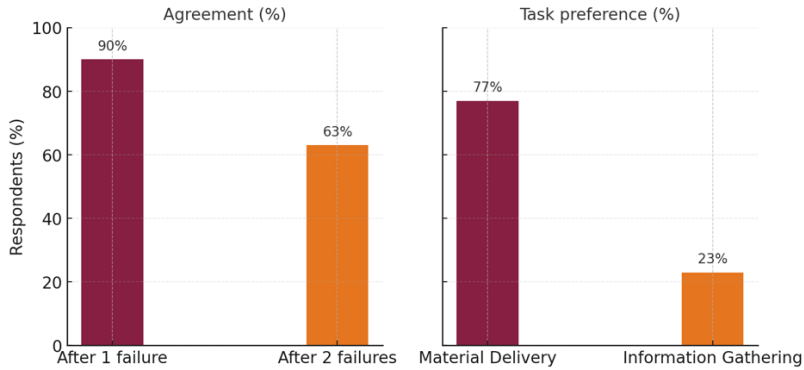

*Figure 14* (left) Percentage of agreement with the second chance to the robot for a new task (right) Preference percentage of users for new reallocation for each tasks

As the reasons behind these choices, those favoring material delivery explained that (1) physical assistance is more demanding for humans, so even imperfect help is valuable. This is aligned with prior studies that removing physical burden is a key for the acceptance of robots in construction industry [71]; (2) a wrong item can be corrected on the next trip if navigation is reliable, so the cost of a single error can be understood; and (3) failures are visible and concrete, which makes it easier to judge progress and decide whether to keep trusting or not. Reviews of construction robotics similarly note that material handling and site logistics are early adoption targets because they combine visible progress with manageable risk [72]. Those preferred information gathering argued that (1) humans also miscount under time pressure, so a counting error is understandable. Therefore, future researchers and practitioners can utilize this opinion to add verification steps from humans that not only make the mistakes understandable but also keep human in the loop [73]; (2) information errors are cheaper to verify with a second check, whereas delivery errors can force rework and further costs; and (3) misdelivery can create downstream safety or quality risks if unnoticed, whereas a count mistake is less hazardous. For example, if the worker is in the dangerous situation and rely on the robot to bring him a critical material, these mistakes can cause further safety challenges.

Based on these arguments we can interpret following insights. First, tolerance for occasional failure is higher when the task removes physical burden and when recovery is cheap, which fits many logistics jobs on site (e.g., fetching tie wire, delivering drill bits, shuttling small tools). In these cases, trust can be sustained by fast retries, safe navigation, and graceful recovery behaviors (e.g., automatic return trips, clear confirmation of item type). Second, information tasks need credibility and transparency more than just simple robot expressions for apologies. For counting materials, progress checks, or hazard scans, workers want evidence (photo snippets, confidence scores, time-stamped logs) and easy verification rather than apologies; designing these feedback loops directly into the interface should stabilize trust far more than expressions alone. The reallocation results (left in Figure 14) also suggest trust thresholds that designers can utilize for the future implications. After a single failure, most participants were still willing to try again, but after two failures a large minority stopped trusting robot. In deployment, robots could watch for back-to-back errors and use recovery strategies and behaviors to sustain the users' trust. For example, switch to a verification mode (show your work), request human confirmation before continuing, or route the task to a fallback policy. For delivery jobs, prioritizing reliable path planning and safe handoffs can help to sustain the trust but it's not enough; and for information jobs, prioritize measurement accuracy, and explainable outputs are essential notes to consider for future scalable work. This approach is also consistent with HRI trust-repair studies that find apologies help some users, but explanations and corrective action are often needed to fully restore trust, especially in safety-critical settings like construction [74], [75].

### 5.4. Implications

The purpose of this section is to discuss practical insights from our findings and highlight three main implications for the design and deployment of robots in construction: (1) treating trust as a dynamic, performance-driven factor, (2) aligning deployment strategies with task type and worker profile, and (3) making interactive expressions and integrating trust and learning mechanisms into HRI systems. First, our results showed that trust changes sharply with outcomes: successful performance builds it quickly, failures reduces it even faster, and apologies can partially repair it, through warmth and communication channel, rather than competence-based perspective. This suggests that, on construction sites, practitioners should consider correcting errors while attempting to explain. For physical tasks such as material delivery, trust repair is tied to corrective actions, so systems should be designed with fast retries, safe navigation, clear confirmation of items, and short "try again" loops. These features help preserve workers' trust to allocate tasks even after one or two mistakes. For information-gathering tasks, gaining trust rests more on evidence and transparency than social expressions. Showing an annotated photo, confidence score, or timestamp, and enabling workers to verify with a single click can improve trust resilience. A practical guideline from our findings is to integrate a "error threshold" (e.g., after two failures) in the robot behavior tree, where the system switches into a verification or assist mode and provide explanations of results. These behaviors can also include requesting minimal corrective input, or implementing a fallback policy, so that repeated errors do not impact trust.

Second, early adoption in construction will likely succeed where robot behaviors are dynamically changed with user characteristics, such as age, education level and etc. Younger participants responded strongly to both successes and failures (easy to gain, easy to lose), older participants were more conservative and demanded proof, while mid-20s users showed more stable trust once it was reestablished. Interfaces should reflect these differences along with environmental situations, where on noisy sites, visual expressions should take precedence over voice; expressive styles should be matched carefully to the robot's morphology to avoid negative effects. Third, make workers and engineers active participants in trust building. It is recommended to give the users the authority and control over the robot expressions that shape the robot repair style and evidence level. Additionally, similar methodologies can be utilized in robot adopted construction sites to generate a trust monitoring system alongside safety and productivity metrics. This monitoring system can include success rate by task, consecutive-error rate, reallocation rate after failures, and which repair style users chose and rated as useful. Therefore, all three measures can be established as foundation for practitioners.

### 5.5. Limitation and Future Work

The experiment results showed that expressive behavior can shape trust dynamics on construction-style tasks, but several limitations still exist. First**,** we studied N=30 participants drawn mostly from a single university population with relatively high educational attainment. While this sample is sufficient for detecting medium within-subject effects, it is not representative of the construction workforce. Future work should expand to a stratified, role-balanced sample that includes craft workers, foremen, superintendents, safety managers, engineers, and project managers across multiple firms and regions. This larger, more diverse sample will allow us to examine subgroup effects (e.g., craft vs. management, apprentices vs. journeymen), increase statistical power for interaction terms (age × task type × repair strategy), and test whether the two-failure redelegation threshold and the task-specific preferences we observed hold in field populations. Second**,** we ran the study in a controlled lab mock-up to ensure participant safety and experimental control. Although the layout approximated job-site conditions, it did not capture the full-time pressure, noise, congestion, and competing priorities of active construction. We also evaluated only two common task families (material delivery and information gathering) with simplified failure modes. Future work will move to supervised field trials on live sites, in partnership with safety officers, to test the framework under real constraints (e.g., limited aisle widths, moving equipment, weather, shift changes). Field studies can instrument longer horizons (days to weeks) to measure trust drift and durability of repair, and can include additional tasks such as tool staging, waste runs, progress photo capture, and simple hazard checks. These studies will also let us test operational policies derived from our findings (e.g., automatic escalation after two consecutive errors, verification workflows for information tasks, or gated autonomy for high consequence moves).

Third**,** while self-report is interpretable and widely used, but it can miss rapid, implicit changes. In the future work, it's recommended to harness more methods like psychophysiological signals (eye gaze, heart-rate variability) collected in a privacy-preserving way. We also used a single platform and embodiment (quadruped with a dog-like face) and a minimal behavior tree for repair (a short apology in addition to one retry). Moreover, future work can compare repair families, for example comparing apology-only, explanation-only, fix-first-then-express, confidence displays, and ask-for-help behaviors. In addition, Future work can introduce context factors common on sites, such as time pressure, noise level, supervisor presence, and task interdependence, to quantify how these stressors moderate the size and half-life of trust repair.

### 6. Conclusion

The purpose of this study was to move one step away from static trust ratings by observing how trust changes over time on construction-style tasks and by testing to what extend expressive repair helps after errors. Using a within-subject design, we measured trust at baseline, after success, after a scripted failure, after an apology-based repair, and after a short reflection, across two task families of Material Delivery (physical assistance) and Information Gathering (perceptual assistance). Three patterns consistently emerged. First, performance events dominate trust: success quickly lifts trust; failure causes a sharp drop. Second, apology-based repaired trust partially, mostly from the perspective of communication channels than competence. This means users felt better about the interaction after robot expressions, but this only partly repair beliefs about reliability and capabilities of robot. Third, nature of task and user profiles impacted the dynamic of trust. Delivery and information tasks show different trust dynamics, and age and prior attitudes shape how strongly trust reacts and how long repair lasts.

To achieve, the overall objective of our study, this paper designed three main research questions, and the responses to those questions are discussed in the following sections. **RQ1** (Dynamics): Expressive behaviors repaired trust but did not replace performance. Trust rose from a baseline of 65.3 to 76.5 after a successful delivery with a glad expression, then fell to 57.0 after failure; the same rise and fall pattern appeared in information gathering (success 74.2, failure 54.1). Robot expressions repaired the lost trust in both tasks (mostly in task delivery category and partially in information gathering task). **RQ2** (Repair effectiveness): The sad-face apology recovered 44% of the trust lost in Material Delivery (to 65.6 and bring it back to the baseline) and 38% in Information Gathering (to 61.8 four points below the baseline). Item-level analysis showed strong recovery in communication and interaction channel, partial recovery in competence channel, and modest repairs in autonomy/decision-making, indicating that apology alone is not sufficient where credibility depends on the actual work. Reallocation behavior of users supported this by showing the willingness to "try again" stayed high after one failure (90%) but dropped after two (63%). **RQ3** (Moderators): Task nature moderated both the size of drops and the durability of repair. Delivery produced larger gains after success, smaller losses after failure, and more effective repair than information tasks. Age moderated sensitivity and retention of trust, where 18-24 reacted strongly in both directions and lost repaired trust quickly while 25-30 showed calibrated, durable repair, and beyond 30 group were conservative, smaller boosts after success, clear penalties after failure, and little drift after reflection. Moreover, the study found that prior attitudes impacted the trust. Optimistic participants buffered single errors and were accepted the failure as a nature of human robot interaction and largely indifferent to apologies without asking for new evidence, while pessimistic participants weighted observable correction and action over social expressions as the most important factor.

For future practice, the results suggest three actionable lessons. (1) Incorporate corrective actions as part of repair strategies with robot expressions. On delivery tasks, trust is remained as the robot shows the capability of recovery is through safe navigation, and pick up. On information tasks, trust repair showed the correlation with transparency and evidence based of expression, where higher trust repair was achievable if the robot incorporate transparency and evidence, such as showing the photo workspace, include a confidence score for its decision, and allow an interactive system that users can verify the results with interaction. (2) adapt trust calibration strategies to user and context. Younger workers may benefit from richer, but brief expressive signals immediately after events; older workers respond better to proof (explanation and evidences of actual works).

Match the modality to the morphology of robot and to site acoustics (prioritize visual signals when noise is high). (3) Integrating error thresholds in robot's behavior tree. Because willingness to reallocate drops after repeated errors (even two errors), integrating error triggers that automatically switch to verification mode, request human confirmation, or route to a safer fallback can enhance the human robot trust. These policies turn trust from an abstract attitude into a managed, auditable variable tied to task risk. Despite these valuable findings, this study is not an exception from limitations, including limited simulated items in lab setting, the size of student sample, covering two task families not more, and facial expression and apology-centric repair strategy. Future studies should test longer deployments on live sites, compare repair strategies (apology vs. explanation vs. fix-first), increase participant pools to reflect craft and supervisory roles, and integrate lightweight behavioral and physiological markers to complement self-reports.

# Appendix A.
**Demographic information:**

**Age:** "What is your age?"
**Gender:** "What is your gender?"
Options:
Male, Female, Non-binary, Prefer not to say, Other (please specify): __________
**Education Level:** "What is the highest level of education you have completed?"
Options:
Bachelor's degree, Master's, PhD, Prefer not to say
**Prior Experience with Robots:** "Have you had any prior experience interacting with robots?"
Options:
Yes
No

**General Attitudes Towards Robots Scale:**

**Instructions:**
For each statement, indicate how often you believe the robot will exhibit the described behavior or trait. Select the response that best reflects your opinion.

| Likert Option |
| --- |
| 1 - Strongly Disagree |
| 2 - Disagree |
| 3 - Neutral |
| 4 - Agree |
| 5 - Strongly Agree |

**Questions:**

1. Robots are necessary because they can do jobs that are too hard or too dangerous for people.
2. Robots can make my life easier
3. Robots are a natural product of our civilization.
4. Widespread use of robots is going to take away jobs from people
5. If I became too dependent on robots, something bad might happen.
6. I would feel very nervous just being around a robot.
7. Robots scare me.
8. Robots can be trusted.
9. Professionally supervised robots are safe enough to use in construction tasks.
10. I would feel relaxed talking with robots.

**Appendix B.**

**Instructions:**
For each statement, indicate **how often** you believe the robot will exhibit the described behavior or trait. Select the response that best reflects your opinion, where each option corresponds to a percentage range:

| Likert Option | Percentage Range |
|---|---|
| 1 – Strongly Disagree | 0–20% of the time |
| 2 – Disagree | 21–40% of the time |
| 3 – Neutral | 41–60% of the time |
| 4 – Agree | 61–80% of the time |
| 5 – Strongly Agree | 81–100% of the time |

**Example:**
You will be asked to rate how often you think the robot will show certain behaviors or qualities.

"The robot is reliable" → You select 4-Agree

**What this means:**

Based on your understanding or experience, you believe the robot behaves in a reliable way **61–80% of the time.**

**Questions:**

1. The robot is **considered a part of team**.
2. The robot is **reliable**.
3. The robot behaves in a **predictable** manner.
4. The robot acts in a **supportive** way.
5. The robot appears **friendly**.
6. The robot acts **autonomously**.
7. The robot **meets the needs of task.**
8. The robot is **pleasant** to interact with.
9. The robot will act **responsibly**.
10. The robot **provides appropriate feedback**.
11. The robot **communicates** with you in an effective manner.
12. The robot **acts consistently**.
13. The robot acts **responsibly**.
14. The robot makes **adequate decisions** on its own.

**Appendix C.**

## Document 2: Wall Specification

1- **WHAT IS MY TASK:** Your task is to build a Wall with foam bricks and blocks available for you. Your finished work should like the figure below.

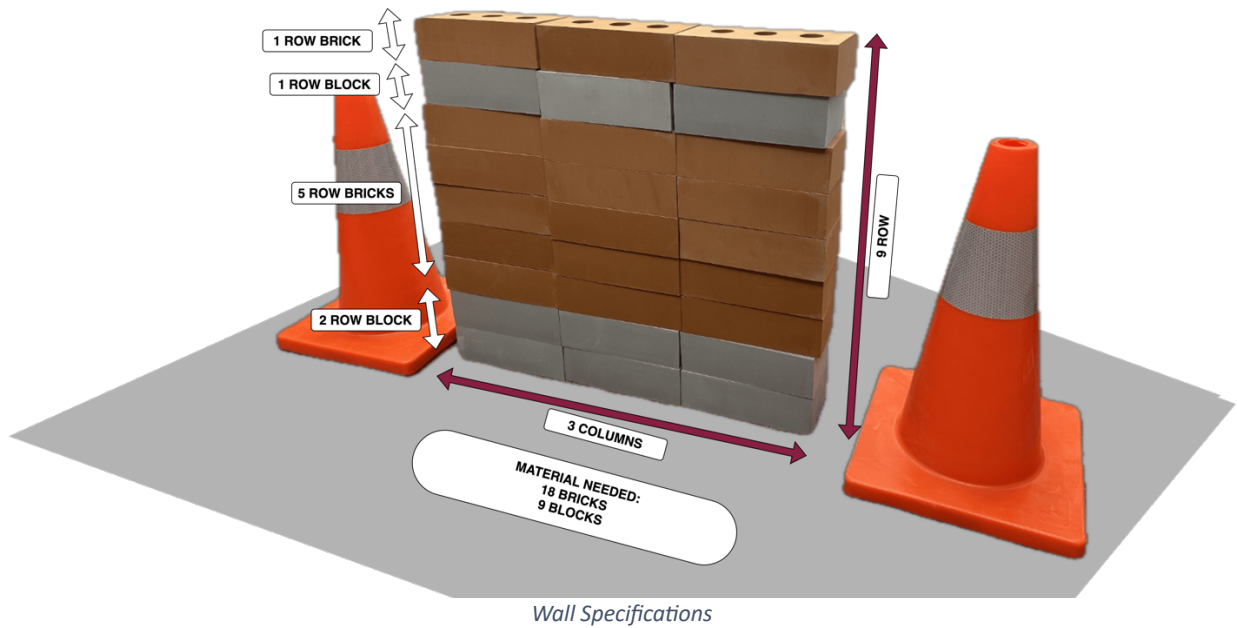

*Wall Specifications*

2- **IS YOUR MATERIAL ENOUGH?** Please check your material and see how many more material you need. (HINT: Some Blocks are not available. How many? )

3- **COLLABORATE WITH ROBOT:** Ask ROBO-1 to bring the required material for you.

4- **HOW TO COLLABORATE:** Use the given Laptop and request your required material using the design button. If you click the button one time, the robot will bring 1 Block/Brick for you.

[27] H. Liu and T. Hiraoka, "Driving behavior model considering driver's over-trust in driving automation system," *Adjunct Proceedings - 11th International ACM Conference on Automotive User Interfaces and Interactive Vehicular Applications, AutomotiveUI 2019*, pp. 115–119, Sep. 2019, doi: 10.1145/3349263.3351525.

[28] Y. Yamani, S. K. Long, and M. Itoh, "Human-Automation Trust to Technologies for Naïve Users Amidst and Following the COVID-19 Pandemic," *Hum Factors*, vol. 62, no. 7, pp. 1087–1094, Nov. 2020, doi: 10.1177/0018720820948981.

[29] P. Robinette, A. M. Howard, and A. R. Wagner, "Effect of Robot Performance on Human-Robot Trust in Time-Critical Situations," *IEEE Trans Hum Mach Syst*, vol. 47, no. 4, pp. 425–436, Aug. 2017, doi: 10.1109/THMS.2017.2648849.

[30] R. Gervasi, L. Mastrogiacomo, and F. Franceschini, "A conceptual framework to evaluate human-robot collaboration," *International Journal of Advanced Manufacturing Technology*, vol. 108, no. 3, pp. 841–865, May 2020, doi: 10.1007/S00170-020-05363-1/FIGURES/11.

[31] M. Chen, H. Soh, D. Hsu, S. Nikolaidis, and S. Srinivasa, "Trust-Aware Decision Making for Human-Robot Collaboration," *ACM Transactions on Human-Robot Interaction (THRI)*, vol. 9, no. 2, Jan. 2018, doi: 10.1145/3359616.

[32] M. Natarajan *et al.*, "Human-Robot Teaming: Grand Challenges," *Current Robotics Reports 2023 4:3*, vol. 4, no. 3, pp. 81–100, Aug. 2023, doi: 10.1007/S43154-023-00103-1.

[33] S. Patil, V. Vasu, and K. V. S. Srinadh, "Advances and perspectives in collaborative robotics: a review of key technologies and emerging trends," *Discover Mechanical Engineering 2023 2:1*, vol. 2, no. 1, pp. 1–19, Aug. 2023, doi: 10.1007/S44245-023-00021-8.

[34] M. Sridharan and B. Meadows, "Towards a Theory of Explanations for Human–Robot Collaboration," *KI - Kunstliche Intelligenz*, vol. 33, no. 4, pp. 331–342, Dec. 2019, doi: 10.1007/S13218-019-00616-Y/TABLES/1.

[35] G. Charalambous, S. Fletcher, and P. Webb, "The Development of a Scale to Evaluate Trust in Industrial Human-robot Collaboration," *Int J Soc Robot*, vol. 8, no. 2, pp. 193–209, Apr. 2016, doi: 10.1007/S12369-015-0333-8/METRICS.

[36] S. You, J. H. Kim, S. H. Lee, V. Kamat, and L. P. Robert, "Enhancing perceived safety in human–robot collaborative construction using immersive virtual environments," *Autom Constr*, vol. 96, pp. 161–170, Dec. 2018, doi: 10.1016/J.AUTCON.2018.09.008.

[37] W. C. Chang, S. M. Ryan, S. Hasanzadeh, and B. Esmaeili, "Attributing responsibility for performance failure on worker-robot trust in construction collaborative tasks," *Proceedings of the European Conference on Computing in Construction*, vol. 4, pp. 0–0, Jul. 2023, doi: 10.35490/EC3.2023.205.

[38] P. Robinette, W. Li, R. Allen, A. M. Howard, and A. R. Wagner, "Overtrust of robots in emergency evacuation scenarios," *ACM/IEEE International Conference on Human-Robot Interaction*, vol. 2016-April, pp. 101–108, Apr. 2016, doi: 10.1109/HRI.2016.7451740.

[39] M. Salem, G. Lakatos, F. Amirabdollahian, and K. Dautenhahn, "Would You Trust a (Faulty) Robot?," *Proceedings of the Tenth Annual ACM/IEEE International Conference on Human-Robot Interaction*, vol. 2015-March, pp. 141–148, Mar. 2015, doi: 10.1145/2696454.2696497.
31